\documentclass{article} 
\usepackage[dvipsnames]{xcolor}

\usepackage[final]{ml4eng_2020}

\usepackage{caption}
\usepackage{subcaption}
\usepackage{hyperref}
\usepackage{url}
\usepackage{graphicx}
\usepackage{multirow}
\usepackage{cancel}
\usepackage{amsmath,bm}
\usepackage{amsfonts}
\usepackage{algorithm}
\usepackage[noend]{algpseudocode}
\usepackage{nicefrac}
\usepackage{mathtools}
\usepackage{multicol}
\usepackage{booktabs}
\usepackage{tikz}
\usetikzlibrary{shapes.geometric}
\usetikzlibrary{shapes.arrows}
\usetikzlibrary{shapes,fit}
\usepackage{array}
\usepackage{makecell}
\usepackage[colorinlistoftodos]{todonotes}

\newtheorem{thm}{Theorem}

\DeclareMathOperator*{\argmin}{argmin}

\usepackage{wrapfig}
\usepackage{etoolbox}

\title{Differentiable Implicit Layers}

\author{%
	Andreas Look$^1$, Simona Doneva$^2$, Melih Kandemir$^1$ , Rainer Gemulla$^2$, Jan Peters$^{3}$, 
	\And
	$^1$Bosch Center for Artificial Intelligence\\
	Renningen, Germany\\
	\texttt{\{andreas.look, melih.kandemir\}@bosch.com} \\
	\And
	$^2$Data and Web Science Group\\
	University Mannheim, Germany\\
	\texttt{\{sdoneva, rgemulla\}@uni-mannheim.de} 
	\And
	$^3$Intelligent Autonomous Systems\\
	TU Darmstadt, Germany\\
	\texttt{peters@ias.tu-darmstadt.de} 
}

\begin{document}

\maketitle

\begin{abstract}
In this paper, we introduce an efficient backpropagation scheme for non-constrained implicit functions. These functions are parametrized by a set of learnable weights and may optionally depend on some input; making them perfectly suitable as a learnable layer in a neural network. We demonstrate our scheme on  different applications: (i) neural ODEs with the implicit Euler method, and (ii) system identification in model predictive control. 
\end{abstract}
\begin{wrapfigure}{r}{0.44\textwidth}
\vspace{-1.0cm}%
    \begin{center}
		\includegraphics[height=.33\columnwidth]{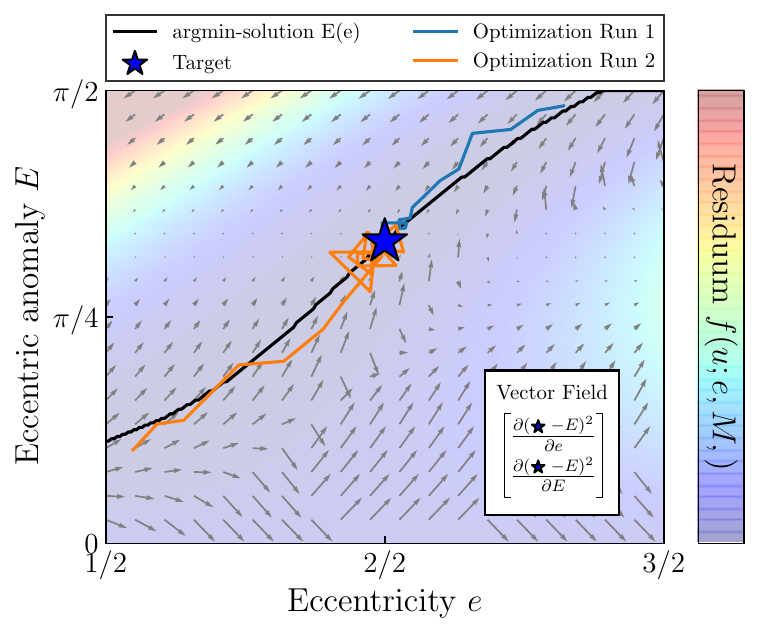}
		\caption{Kepler's Equation implicitly defines $E$ as a function of $e$ for a given $M$. 
		We optimize $e$ such that the implicitly defined  $E(e)$ matches a target value. 
		The IFT provides the means for estimating the implicitly defined gradient $\nicefrac{\partial E}{\partial e}$.
		The gradient field provides information even at inexact $\argmin$-solutions, i.e. points that are not on the $\argmin$-line also point to the target.}
		\label{fig:1}
		\vspace{-5pt}
	\end{center}
\end{wrapfigure}

\section{Introduction}
\label{sec:intro}

Implicit functions can be found in a wide range of domains, e.g. physics, numerics, or math. A famous example is  Kepler's equation: $M = E - e \sin(E)$, which is elemental in orbital mechanics (see Fig \ref{fig:1}). It estimates the relation between the eccentric anomaly $E$,  mean anomaly $M$, and eccentricity $e$. Contrarily, learning such an implicitly defined function is not feasible with the standard deep learning practice, since it commonly consists of a chain of functional mappings described by algebraic operations. We introduce the framework of {unconstrained} and {non-convex} \textit{Differentiable Implicit Layers} (DIL) as a plug-and-play extension for neural networks that enables efficient learning of such implicitly defined problems. An implicit layer \citep{gould2019deep} is defined as a mapping that takes an input $\bm x \in \mathbb{R}^{D_x}$ and produces an output $\bm y \in \mathbb{R}^{D_y}$  that is obtained as an $\argmin$-solution to the scalar-valued score function  $f:\mathbb{R}^{D_y + D_x + D_\theta} \rightarrow \mathbb{R}$, parameterized by $\bm \theta \in \mathbb{R}^{D_\theta}$:
\begin{align}
    \bm y := \argmin_u f(\bm u;\bm x, \bm \theta).
    \label{eq:param_argmin}
\end{align}
We can interpret Kepler's equation as an $\argmin$-problem with the parameters $e$ and $M$: $E=\argmin_u (u - e \sin(u)-M; e, M)^2$. 
During network training, we target to optimize the parameters $\bm \theta$, such that the output $\bm y$ of the implicit layer exhibits a desired behaviour on a subsequent task, i.e. minimizes a scalar loss $\mathcal{L}(\cdot)$. 
Consequently, we need to solve the nested $\argmin$-problem:
\begin{align}
    \argmin_{ \theta} \mathcal{L} \Big (  \argmin_u f(\bm u;\bm x, \bm \theta), \bm x, \bm \theta   \Big ).
    \label{eq:param_argmin_outer}
\end{align}
Note that the loss function may also depend on the parameters $\bm \theta$ (acting as a regularizer) and the input $\bm x$. 
Likewise, we aim to estimate in Fig. \ref{fig:1} the correct $e$ such that the implicitly defined eccentric anomaly $E$ matches a target value.

Most research on implicit layers for neural networks focuses on specific architectures \citep{liao2018reviving, deepeq} or $\argmin$-problem classes \citep{optnet, diff_cem}, e.g. of convex type \citep{amos_convex}. Concurrent work on general implicit networks \citep{gould2019deep, zhang2020implicitly} without any restriction on problem or network type, did not scale to training of heavily parameterized implicit layers with high dimensional output. 
Common handicap of the aforementioned general approaches is the explicit calculation and inversion of large Jacobians. However, the existing solutions are prohibitively costly to be presented as a general purpose layer for neural networks


We propose a method that generalizes existing problem-specific solutions to a more comprehensive framework, while bringing them an unprecedented level of scalability. Our differentiable implicit layer consists of two parts: (i) the learnable $\argmin$-problem, and (ii) the solver. The solver is used only during the forward evaluation, i.e. it does not influence the backward evaluation, by-passing a large set of potential numerical difficulties. 
During training, the solution $\bm y$ is evaluated on the downstream scalar loss function $\mathcal{L}(\cdot)$, for which we provide an efficient backward evaluation scheme by combining the \textit{Implicit Function Theorem} (IFT) and the \textit{Conjugate Gradient Method} (CG). In contrast to prior art, our approach omits the explicit calculation and inversion of large Jacobians, which are typically necessary for IFT evaluation. Our backward evaluation relies solely on 
efficient to estimate \textit{vector-Jacobian products} (VJP).
We summarize our contribution as below:
\begin{itemize}
    \item We propose unconstrained and non-convex parameterized differentiable implicit layers for neural networks as a construct that vastly enhances the feasible problem set for the automatic differentiation technology.
    \item We make differentiable implicit layer training scalable for over-parameterized neural networks with a large output dimensionality.
    \item We demonstrate the efficiency of our method by applying it  to 
    (i) implicit solvers for neural ODEs, and (ii) model predictive control.
\end{itemize}
\section{The proposed Framework}
\label{sec:methodology}
The forward evaluation of a DIL consists of applying a potentially non-differentiable solver to an $\argmin$-problem in order to solve for $\bm y$ by minimizing the score function $f(\cdot~; \bm x, \bm \theta)$. 
The solver is used solely for the forward evaluation.
Hence, we can treat the solver in our proposed framework as a blackbox.
However, the main difficulty in developing an efficient framework for differentiable implicit layers lies in the backward evaluation. When $\bm y$ is passed on to a subsequent task, i.e. a scalar loss function $\mathcal{L}(\cdot)$, the Bi-Level IFT  (Thm. \ref{thm_biift}), which is an extension to the standard IFT (see Appx. \ref{app:ift}), provides an estimate to the gradients $\nicefrac{d \mathcal{L}(\bm y)}{d \bm x}$, and $\nicefrac{d \mathcal{L}(\bm y)}{d \bm \theta}$.
\begin{thm}\label{thm_biift} \textbf{(Bi-Level IFT.)}
Let $\bm y$ be the solution to a parameterized $\argmin$-problem (Eq. \ref{eq:param_argmin}). If $\bm y$ is evaluated on a downstream scalar loss function $\mathcal{L}(\bm y, \bm x, \bm \theta)$ (Eq. \ref{eq:param_argmin_outer}), the gradients with respect to the input $\bm x$ (exchangeable $\bm \theta$) are obtained exclusively by vector-Matrix products as: 
\begin{equation*}
    \frac{d \mathcal{L}}{d \bm x}^T = -
    \underbrace{\frac{\partial  \mathcal{L}}{\partial \bm y}^T
    \overbrace{\left(\frac{\partial^2 f}{\partial \bm y^2} \right)^{-1}}^{\bm H^{-1}}}_
    {\text{vector-inv. Hessian product $\coloneqq     g^T$}}
    \left( \frac{\partial^2 f}{\partial \bm x \partial \bm y}  \right)+ \frac{\partial \mathcal{L}}{\partial \bm x}^T = -
    \underbrace{\bm g^T \left( \frac{\partial^2 f}{\partial \bm x \partial \bm y}  \right)}_
    {\text{VJP}}+ \frac{\partial \mathcal{L}}{\partial \bm x}^T.
\end{equation*}
\end{thm}
\paragraph{Conjugate-Gradient-Method.} Explicitly inverting the Hessian $\bm H$ is intractable during training of a neural network, since it is computational too expensive $\propto \mathcal{O} ( {D_y}^3)$. 
Moreover modern automatic differentiation libraries lack the capability of estimating the Hessian efficiently. 
Instead, we directly estimate the vector-inverse Hessian product $\bm g$ as a solution to the \textit{linear system of equations} (LSE):
\begin{equation}
    \label{eq:cg_lse}
    \underbrace{\bm H \overbrace{\left(\bm H^{-1}\frac{\partial  \mathcal{L}}
    {\partial \bm y} \right)}^{\bm g}}_{\textit{VJP: } (\bm g^T \bm H)^T  } = \frac{\partial \mathcal{L}}{\partial \bm y}.
\end{equation}
Since the Hessian is evaluated at a minimum, i.e. the solution $\bm y$ to the $\argmin$-problem, the Hessian $\bm H$ is \textit{positive semi-definite} (PSD) and 
the conjugate gradient method is suitable for solving the LSE. 
The resulting LSE can be solved via the CG method without the need of evaluating the Hessian explicitly.  
Each CG step requires one $\texttt{grad}$-function call, which estimates the  \textit{vector-Jacobian product} (VJP),
and converges in the absence of round-off errors after at most $D_y$ steps \citep{iterative_method}.
In contrast, the naive method of explicitly inverting the Hessian $\bm H$ requires firstly $D_y$ VJP evaluations in order to build the Hessian, 
which are as many as CG requires for the full evaluation of the vector-inverse Hessian product $\bm g$. 
The costly inversion of the Hessian comes additionally on top.

\paragraph{Algorithm.}
We summarize our framework in Alg. \ref{alg:main}. During the $\texttt{forward}$-evaluation of a DIL the score function $f$ with   optional input $\bm x$
is minimized with a blackbox $\texttt{solver}$.
As a result we obtain the solution $\bm y$.
The $\texttt{backward}$-evaluation receives the vector-valued gradient of the loss function $\mathcal{L}(\cdot)$ with respect to the optimal 
solution $\bm y$, i.e. $\nicefrac{\partial \mathcal{L}(\bm y)}{\partial \bm y}$. 
The gradients with respect to the parameters $\bm \theta$ and input $\bm x$ are estimated via the Bi-Level IFT (Thm. \ref{thm_biift}).
The function $\texttt{VJP\_CG}$ uses a CG method, which 
relies on vector-Jacobian products, in order to estimate vector-inverse Hessian product  $\bm g$ without explicitly calculating the
Hessian $\bm H$. 

\begin{algorithm}[ht]
    \scriptsize
	\caption{Forward/ Backward Evaluation for Differentiable Implicit Layers}\label{alg:main}
	\begin{algorithmic}
		\State \textbf{Input:} Score Function $f(\cdot;\bm x, \bm \theta)$, Parameters $\bm \theta$, \texttt{solver}$(\cdot)$
		\Function{\textnormal{\textcolor{blue}{\texttt{Forward}}}}{$\bm x$} \Comment{Optional Input $\bm x$}
		\State $\bm y = \texttt{solver}(f(\cdot;\bm x, \theta))$ \Comment{Solve for $\argmin$-solution $\bm y$ with user defined \texttt{solver}}
		\State \textbf{return}  $\bm y$
		\EndFunction %
		
		\Function{\textnormal{\textcolor{OliveGreen}{\texttt{Backward}}}}{$\bm y, \nicefrac{\partial \mathcal{L}}{\partial \bm y}$} 
		\Comment{According to Thm. \ref{thm_ift}}
		\State Set $\bm g_1 = \nicefrac{\partial f}{\partial \bm y}$ 
		, $\bm g_2 = \nicefrac{\partial \mathcal{L}}{\partial \bm y}$                \Comment{Score, Loss function gradient at optimal solution}
		\State $\bm g =$ \textcolor{red}{\texttt{VJP\_CG}}{$(\bm y, \bm g_1, \bm g_2)$}  \Comment{vector-inv. Hessian product 
		$ \bm g^T = \frac{\partial \mathcal{L}}{\partial \bm y}^T(\frac{\partial^2 f}{\partial \bm y^2} )^{-1}$}
		\State $\nicefrac{d \mathcal{L}}{d \bm x} = -\texttt{grad}(\bm g_1, \bm x, \texttt{grad\_outputs}=\bm g)^T $ 
		\Comment{Return $-\bm g^T \frac{\partial^2 f}{\partial \bm x \partial \bm y}$ in $\frac{d \mathcal{L}}{d \bm x}^T$}
		\State $\nicefrac{d \mathcal{L}}{d \bm \theta} = -\texttt{grad}(\bm g_1, \bm \theta, \texttt{grad\_outputs}=\bm g)^T $
		\Comment{Return $-\bm g^T \frac{\partial^2 f}{\partial \bm \theta \partial \bm y}$ in $\frac{d \mathcal{L}}{d \bm \theta}^T$}
		\State \textbf{return}  $\nicefrac{d \mathcal{L}}{d \bm x}, \nicefrac{d \mathcal{L}}{d \bm \theta}  $
		\EndFunction %
		
		\Function{\textnormal{\textcolor{red}{\texttt{VJP\_CG}}}}{$\bm y, \nicefrac{\partial f}{\partial \bm y}, \nicefrac{\partial \mathcal{L}}{\partial \bm y}$} 
		\Comment{CG with efficient $\texttt{grad}$ calls} 
		\State Init $\bm x_0$
		\State Set $\bm r_0 = \nicefrac{\partial \mathcal{L}}{\partial \bm y} - (\texttt{grad}(\nicefrac{\partial f}{\partial \bm y}, \bm y, \texttt{grad\_outputs}=\bm x_0)^T+\epsilon \bm x_0$) \Comment{Add $\epsilon \bm x_0$ to tackle a singular Hessian}
		\State Set $\bm p_0 = \bm r_0, k=0$
		\While{$|| \bm r_k || > tol$}
		\State $\textcolor{orange}{\bm A \bm p_k}  = \texttt{grad}(\nicefrac{\partial f}{\partial \bm y}, \bm y, \texttt{grad\_outputs}=\bm p_k )^T +\epsilon \bm p_k$ \Comment{Symmetric Jacobian: $\text{VJP}=\text{JVP}^T$}
		\State $\alpha_k = \nicefrac{(\bm r_k^T \bm r_k)}{(\bm p_k^T \textcolor{orange}{\bm A \bm p_k})}$
		\State $\bm x_{k+1} = \bm x_k + \alpha \bm p_k$
		\State $\bm r_{k+1} = \bm r_k - \alpha \textcolor{orange}{\bm A \bm p_k}$ 
		\State $\beta_k = \nicefrac{(\bm r_{k+1}^T \bm r_{k+1})}{(\bm r_k^T \bm r_k)}$
		\State $\bm p_{k+1} = \bm r_{k+1} + \beta_k\bm p_k$
		\State $k = k +1 $
		\EndWhile
		\State \textbf{return}  $\bm x_{k+1}$ \Comment{Solution $\bm g$ in $ \bm H \bm g = \frac{\partial \mathcal{L}}{\partial \bm y}$}
		\EndFunction %
\end{algorithmic}
\end{algorithm}

\section{Applications}
\label{sec:experiments}

If not explicitly stated otherwise, we use CG during the evaluation of the IFT (as in Alg. \ref{alg:main}). 
In the first experiment we introduce implicit neural ODEs and compare our method to the adjoint training method
\citep{neural_ode}. 
Lastly we explore our method in the context of model predictive control.

\subsection{Solving Neural ODEs with the Implicit Euler Method}
\label{sec:node}
Dynamical systems are commonly described by an ordinary differential equation (ODE). 
The commonplace way to identify a dynamical system by neural networks is the neural ODE (NODE) \citep{neural_ode}. 
NODEs have been observed to introduce implementation challenges. Firstly, the adjoint training method \citep{neural_ode} is well known to cause numerical instabilities due to non-reversibility of the NODE \citep{anode}. Further, the backward evaluation of the adjoint requires an additional computational costly solution to the induced ODE problem. Our DIL framework is capable of addressing all of these points by introducing an implicit NODE formalism. In the following,  we focus for simplicity on the backward Euler solver. 

\paragraph{Backward Euler NODE.}  
When solving a NODE with the 
backward or implicit Euler method \citep{hairer1993solving}, we obtain the update rule:
\begin{equation*}
    d \bm x = \bm h (\bm x ; \bm \theta) dt \xrightarrow[Euler]{Backward} \bm x_{t+1} = \bm x_t + \bm h(\bm x_{t+1}; \bm \theta) \Delta t,
    \label{eq:backwar_euler}
\end{equation*}
with the state  $\bm x \in \mathbb{R}^{D_x}$ and neural dynamical model $\bm h:\mathbb{R}^{D_x} \rightarrow \mathbb{R}^{D_x}$ with parameters $\bm \theta$.
The backward Euler scheme is L-stable \citep{butcher2003numerical} and
has convergence order 1.  
The property of L-stability, which only implicit solvers have, allows to use larger step sizes and, 
above all, making the method suitable for stiff systems. Note the nuance that the backward Euler method uses $\bm h(\bm x_{t+1}; \bm \theta)$ as opposed to the forward Euler, which uses  $\bm h(\bm x_{t}; \bm \theta)$. Solving such an implicit problem can be translated to residual minimization:
\begin{equation}
    \argmin_{x_{t+1}} r(\bm x_{t+1}; \bm x_t, \bm \theta) = \argmin_{x_{t+1}} || \bm x_{t+1} - (\bm x_t + \bm h(\bm x_{t+1}; \bm \theta) \Delta t) ||.
     \label{eq:node_res}
\end{equation}
When viewing the residual $r(\bm x_{t+1}; \bm x_t, \bm \theta)$ as the learnable score function $ f$ with parameters $\bm \theta$ and input $\bm x_t$,
we obtain a DIL and can evaluate the backward pass with our proposed Alg. \ref{alg:main}. Now it remains open how to estimate the solution $\bm x_{t+1}$. We obtain the solution $\bm x_{t+1}$  via fixed-point iteration for non-stiff problems or for stiff problems via the Newton iteration:
\begin{equation*}
\bm x_{t+1}^{(i+1)} =  
\bm x_{t+1}^{(i)} - \underbrace{\bm H_r^{-1}\frac{\partial r}
{\partial {\bm x_{t+1}^{(i)}} }}
_{\text{inv. Hessian-vector product $\coloneqq \bm g_r$}},
\end{equation*}
with the Hessian $\bm H_r$ of $r(\bm x_{t+1}; \bm x_t, \bm \theta)$. \cite{cheap_operations} propose to approximate $\bm H_r$ by its diagonal values or the identity matrix. However it is more favourable to have an exact evaluation procedure, instead of relying on such approximations. Note that $\bm H_r$ is not necessary PSD, unless it is evaluated at the solution $\bm x_{t+1}$. Consequently, the CG method as defined in Alg. \ref{alg:main} is not applicable in order to estimate $\bm g_r$. 
However, we may still use the CG method if we modify the original LSE   \citep{cg_book} by multiplying both sides with $\bm H_r^T$:
\begin{equation*}
    \underbrace{\bm H_r 
    \overbrace{\left(\bm H_r^{-1} \frac{\partial r}{\partial {\bm x_{t+1}^{(i)}}}\right)}^{\bm g_r}}_{\textit{VJP: } (\bm g_r^T \bm H_r)^T = \tilde{\bm g_r}  }
    =
   \frac{\partial r}{\partial {\bm x_{t+1}^{(i)}} }
\xrightarrow[\text{with } \bm H_r^T]{\text{Multiply both sides} }
    \underbrace{ \overbrace{\bm H_r^T \bm H_r}^{PSD} \bm g_r}
    _{\textit{VJP: } ({\tilde{\bm g_r}}^T \bm H_r)^T }=
    \bm H_r^T \frac{\partial r}{\partial {\bm x_{t+1}^{(i)}} }.
\end{equation*}
Note that $\bm H_r^T \bm H_r$ is PSD and hence CG is applicable. Though the  left-hand side  of the modified LSE looks prohibiting at first sight, it can be evaluated efficiently by any $\texttt{autodiff}$-library via two $\texttt{grad}$-evaluations. Consequently, CG can be used with two $\texttt{grad}$-function calls per iteration. The backward evaluation can be performed by the IFT as proposed in Alg. \ref{alg:main} or alternatively with the adjoint method \citep{neural_ode}. 

\paragraph{Runtime Profiles.}
Although a root finding problem (Eq. \ref{eq:node_res}) needs to be solved during the forward evaluation of a NODE with  the backward Euler method, it is still faster than the default adaptive step size solver DOPRI5 \citep{neural_ode}, as shown in  Fig. \ref{fig:node_times_forward}). If viewing the NODE with backward Euler solver as a DIL, we observe during the backward evaluation a significant decrease in the required computation time compared to the
adjoint method (see Fig. \ref{fig:node_times_backward}). Another benefit of the DIL viewpoint is the independence of the backward evaluation time from the NODE stiffness, which tends to increase throughout the training \citep{cheap_operations}. 
\begin{figure}[h!]
	\centering
	\begin{subfigure}[h]{0.48\columnwidth}
		\centering\includegraphics[height=0.5\columnwidth]{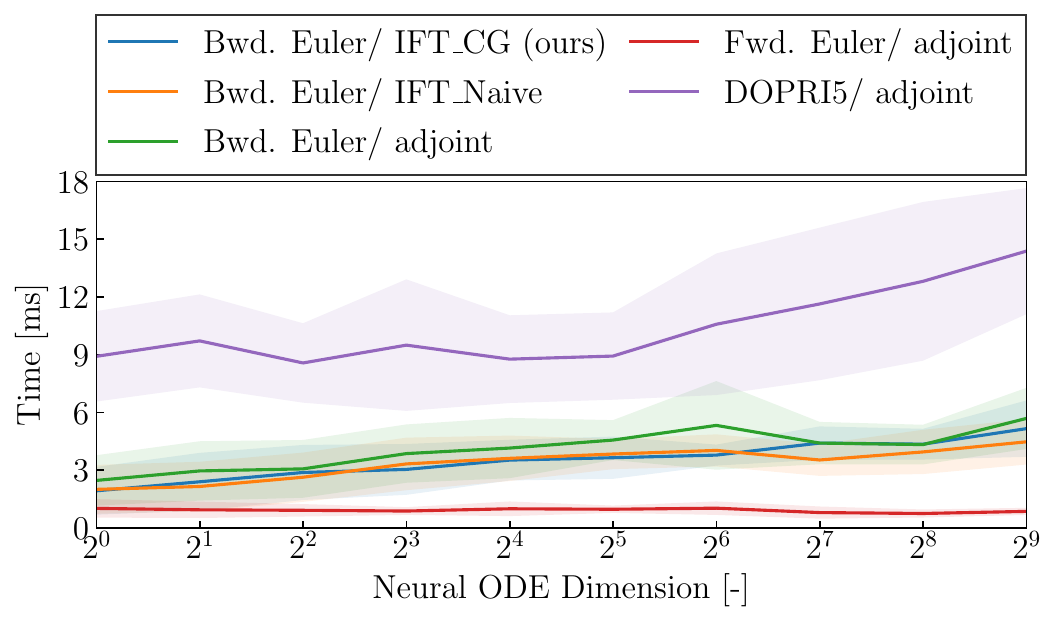}
		\caption{NODE Forward evaluation.}
		\label{fig:node_times_forward}
	\end{subfigure}%
    \hfill
	\begin{subfigure}[h]{0.48\columnwidth}
	\centering\includegraphics[height=0.5\columnwidth]{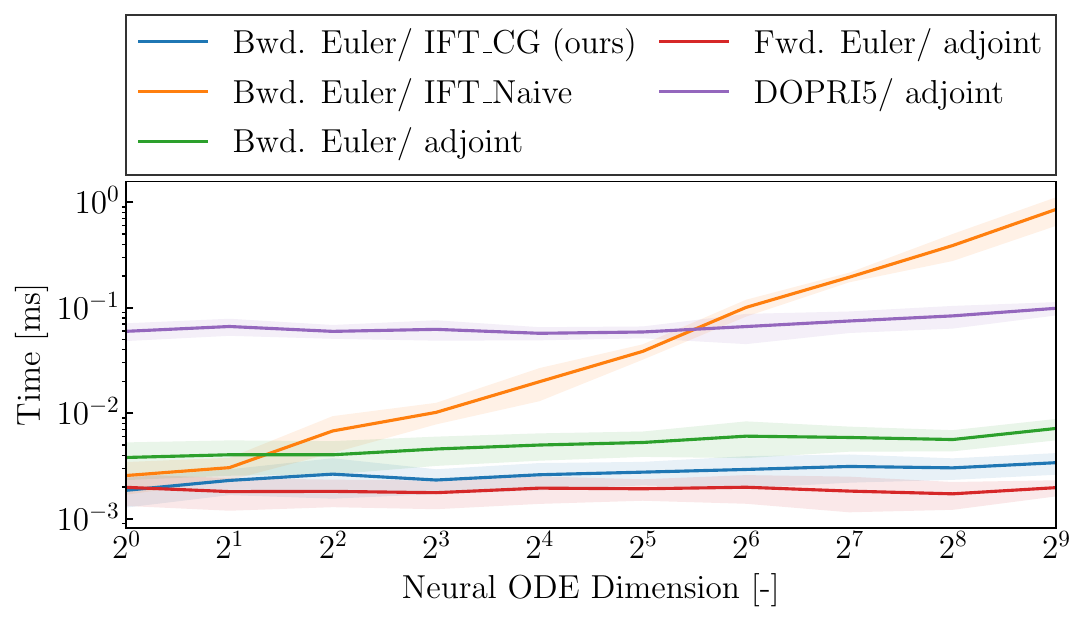}
	\caption{NODE Backward evaluation.}
	\label{fig:node_times_backward}
	\end{subfigure}
	\caption{Mean $\pm$ standard deviation of forward/ backward evaluation times averaged over 100 NODE initializations (Layers: 2, Hidden size: 30). Equal accept criterions of the solution $\bm x_{k+1}$ were used for all methods. 
	}\label{fig:node_times}
\end{figure}

\newpage
\begin{wraptable}{r}{8.0cm}
	\caption{
	Average test MSE and standard errors on two extrapolation tasks: Van der Pol (20runs, 2dim, 106 step extrapolation), Spiral Data (20runs, 2dim, 150 step extrapolation), and CMU Walking (10runs, 50dim, 297 step extrapolation). 
	}
	\label{tab:results}
	\begin{center}
			\scriptsize
			\begin{tabular}{lclc}
				\toprule
				 NODE Models & Van der Pol & Spiral Data & CMU Walking\\
				 \midrule
				 $\text{DOPRI5}_{\text{adj.}}                   $ & 0.89 $\pm$ 0.15 & 0.13 $\pm$ 0.01 & 15.92 $\pm$ 2.10\\
				 $\text{Fwd. Euler}_{\text{adj.}}               $ & 0.68 $\pm$ 0.07  & 0.20 $\pm$ 0.01 & 12.17 $\pm$ 1.39\\
				 $\text{Bwd. Euler}_{\text{adj.}}               $ & 0.67 $\pm$ 0.12 & \bf 0.09 $\pm$ 0.01 & 13.68 $\pm$ 2.02\\
				 $\text{Bwd. Euler}_{\text{IFT, Naive}}  $        & {\bf 0.38 $\pm$ 0.05} & {\bf 0.09 $\pm$ 0.00}  &  {\bf 11.57 $\pm$ 1.79}\\
				 $\text{Bwd. Euler}_{\text{IFT, CG}} \text{(ours)}$ & {\bf 0.40 $\pm$ 0.06} & {\bf 0.09 $\pm$ 0.01} & {\bf 11.43 $\pm$ 1.27}\\
				\bottomrule
			\end{tabular}
	\end{center}
	\vspace{-5mm}
\end{wraptable}
\paragraph{Predictive Performance.}
We benchmark the proposed Backward Euler NODE on three time series forecasting tasks.  
In the first experiment we generate 320 equally spaced observations according to the Van der Pol equation\footnote{$\frac{\partial^2 x}{\partial t^2}-\mu (1-x^2)\frac{\partial x}{\partial t}+x=0$,  with $dt=0.1\,\text{s}$ and $\mu=3$}.
First 107 observations are used for training, next 106 observations for validation, and last 106 observations for testing. 
In the second experiment, we generate 300 equally spaced observations according to spiral dynamics
\footnote{$\frac{\partial \bm x}{\partial t} = \bm A \bm x^3 $,  with $dt=0.1\,\text{s}$ and $a_{1,1}=-1,a_{1,2}=2,a_{2,1}=-2,a_{2,2}=-1/10,$}. 
We use the first 100 points for training, next 50 for validation and the final 150 for testing.
In the third experiment, we follow \cite{odevae} for designing the experimental setup using data from the CMU motion capture library. 
The dataset is split into 16 sequences for training, three for validation, and four for test.  A detailed sketch of the used architectures are given in Appx. \ref{app:archtitect_benode}.  
We observe a consistent performance improvement compared to the adjoint method if the NODE discretized by backward Euler is trained with the IFT.  
Using the much faster CG method during the backward evaluation comes with no performance loss
compared to the naive IFT evaluation.

\subsection{Differentiable Path Planning }
\label{sec:control}

We adapt the well established setup of \textit{model predictive control} (MPC) with moving horizon \citep{Diehl11numericaloptimal}. 
At each time step we observe
the current state of the system  $\bm x_{obs.}$ and plan the optimal trajectory on a limited horizon $H$. 
After planning, the first control $\bm u_0$ is executed and the time step is moved one step forwards.
The optimization problem at the planning step can be formalized as:
\begin{equation}
    \argmin_{u_{0:H}} \sum_{t=0}^{H} c(\bm x_t, \bm u_t; \bm \theta_c) ~~, s.t.  
    ~\bm x_{t+1} = \bm h (\bm x_{t}, \bm u_{t}; \bm \theta_h), ~\bm x_0 = \bm x_{obs.},
    \label{eq:control_problem}
\end{equation}
with the control $\bm u_t \in \mathbb{R}^{D_u}$, state $\bm x_t \in \mathbb{R}^{D_x}$, dynamics $\bm h: \mathbb{R}^{D_x} \rightarrow \mathbb{R}^{D_x}$ with parameters $\bm \theta_h$, and cost function $c:\mathbb{R}^{D_x+D_u} \rightarrow \mathbb{R}$ with parameters $\bm \theta_c$. 
By inserting the constraints we can interpret the optimization problem (Eq. \ref{eq:control_problem}) as an instance of our framework DIL.
We treat the observed state $\bm x_{obs.}$ as the optional input and $\bm \theta_h$, $\bm \theta_c$ as the parameters of the score function.
The output of this implicit layer is the control sequence $\bm u_{0:H}$.
We can efficiently return the derivatives of the control sequence with respect to $\bm x_k$, $\bm \theta_c$, and $\bm \theta_h$  via Alg. \ref{alg:main}.
An alternative approach is obtained by linearizing the optimization problem. 
However, due to ill convergence properties, this approach did not scale to neural dynamical models \citep{diffmpc}. 
In the following paragraph we provide a proof of concept that the cost function can be indeed learned by backpropagation through 
the trajectory planning step, when dynamics is governed by a neural network.
Therefore, we interpret Eq. \ref{eq:control_problem} as a DIL ($\text{MPC}_\text{IFT}$).  
We provide in Appx. \ref{app:lqr} an additional experiment for the case of linear dynamics and cost, in which we recover true dynamics and cost using only the observed 
control sequence. 

\begin{figure}[h]
	\centering
	\begin{subfigure}{0.3\columnwidth}
	\centering\includegraphics[height=0.7\columnwidth]{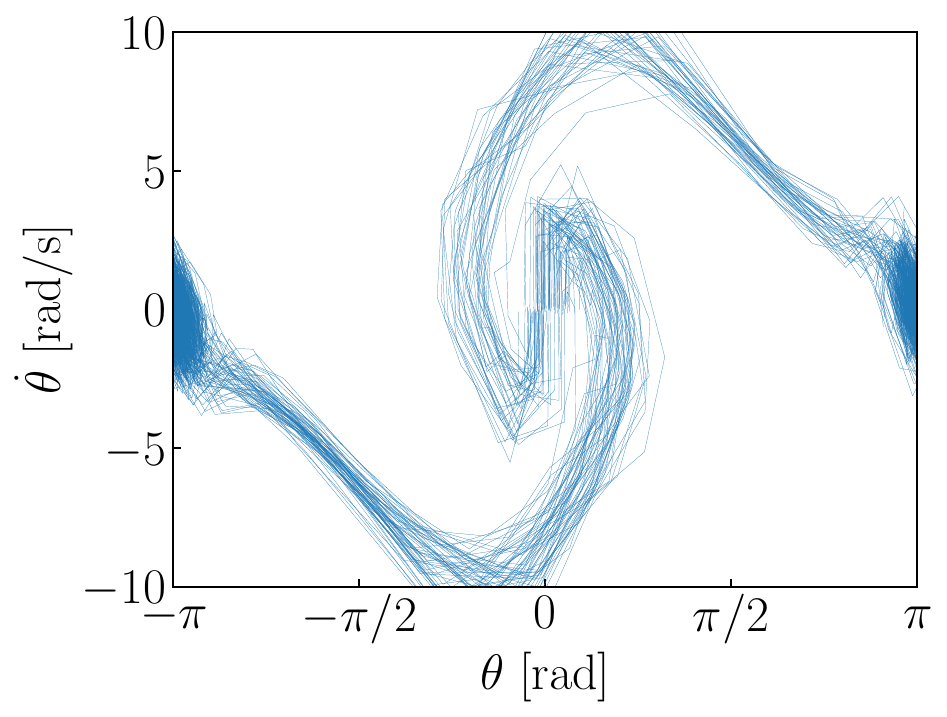}
	\caption{Training Data with low variance at initial state. }
	\label{fig:control_exp}
	\end{subfigure}%
   \hfill
	\begin{subfigure}{0.3\columnwidth}
	\centering\includegraphics[height=0.7\columnwidth]{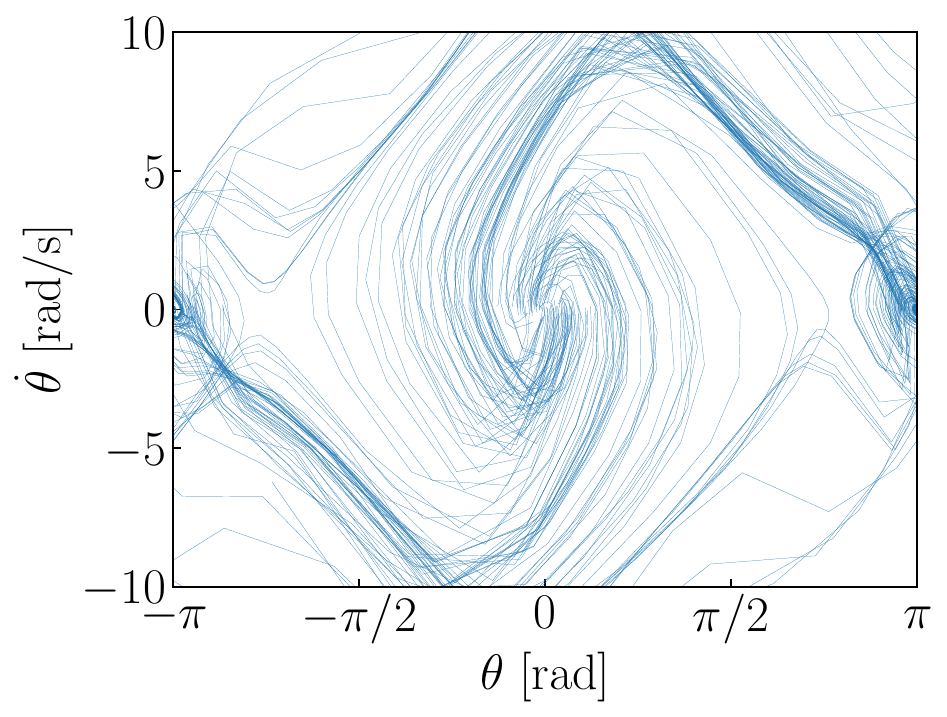}
	\caption{Behavioural cloning with high variance at initial state. }
	\label{fig:control_bco}
	\end{subfigure}%
	\hfill
	\begin{subfigure}{0.3\columnwidth}
	\centering\includegraphics[height=0.7\columnwidth]{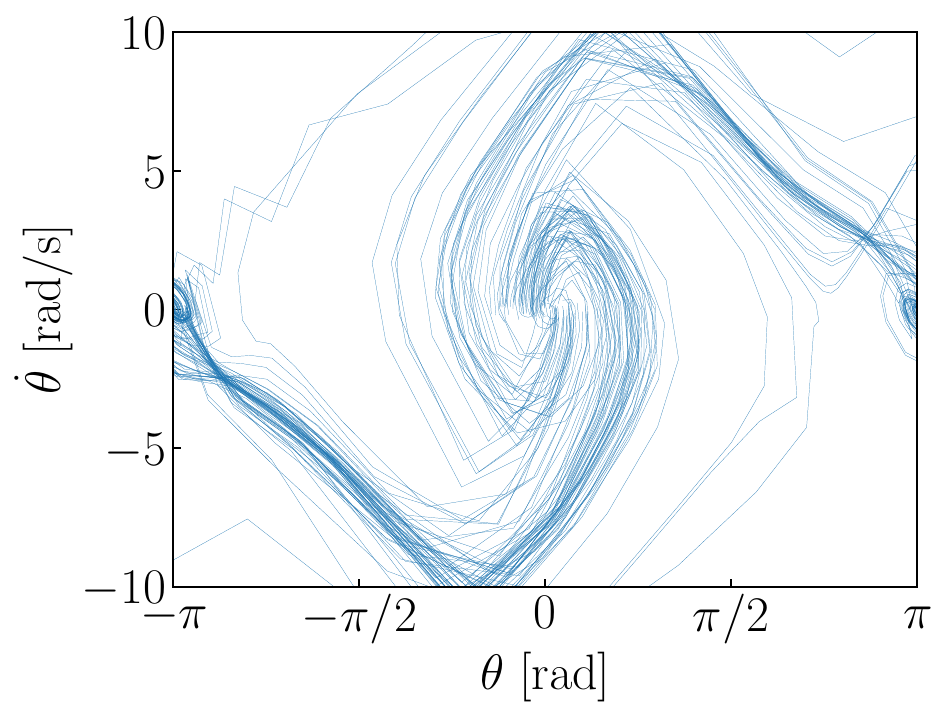}
	\caption{$\text{MPC}_{\text{IFT}}$ (ours) with high variance at initial state. }
	\label{fig:control_mpc}
	\end{subfigure}%
	\caption{Cart pole swing-up trajectories. Ground truth from the expert (left) and learned from expert observations by behavioral cloning (middle) and our method (right).}
	\label{fig:control}
\end{figure}

\paragraph{Imitation Learning from Observations.}
Suppose we observe a dataset $\mathcal{D}_{exp.}$, which consists of $N$ trajectories $\bm x_{1:T}^{1:N}$ with horizon $T$, generated by an expert policy. Note , the controls $\bm u_{1:T}^{1:N}$ are not observed. Let the expert policy be realized as the solution to the MPC problem as defined in Eq. \ref{eq:control_problem}. We target to recover the expert policy by fitting a student policy to a sequence of state transitions observed from the expert \citep{bco, gailfo}. The student policy is also evaluated as the $\argmin$-solution to the MPC problem (Eq. \ref{eq:control_problem}), though with a learned cost and dynamics function. 
We approximate the dynamics functions with neural networks without using any prior information. 
The cost function evaluates the distance between the observed state and a learnable target state.
In our setup we can query the true dynamical model, but do not know its functional form. The learnable dynamics function is trained on $(\bm x, \bm u, \bm x')$ triplets, with $\bm x \sim \mathcal{D}_{exp.}$, $\bm u \sim \mathcal{U}(\bm u_{min}, \bm u_{max})$, and $\bm x'$ as the true next state. The cost function is trained on the MSE between observed expert trajectories and predicted trajectories. 
Alg. \ref{alg:mpc} in Appx. \ref{app:diff_mpc} summarizes the learning procedure.

\paragraph{Imitating an Noisy Expert.} 
\begin{wraptable}[12]{r}{7.2cm}
    \vspace{-4mm}
   	\caption{
	Average cost and standard error for cartpole swingup task (50 initial positions, 10 runs). 
	We test generalization capabilities by testing on  higher variance at the initial state.
	$*$ \citep{Bain96aframework}}
	\label{tab:results_control}
	\small
	\begin{center}
			\scriptsize
			\begin{tabular}{lcc}
				\toprule
				&{Low Variance} &{High Variance}\\
			 Model	& {$x_0 \sim \mathcal{N}(0, 0.04\bm I)$} &{$ x_0 \sim \mathcal{N}(0, 0.08\bm I)$}\\
				 \midrule
				 $\text{Expert}$     & 9.2 $\pm $ 0.0 &  (9.3 $\pm $ 0.1)\\
				 $\text{BC}^{*}$               & 14.6 $\pm$ 0.6        &  18.4 $\pm$ 1.4\\
				 $\text{MPC}_\text{IFT}$(ours) & {\bf   13.7 $\pm$ 0.4}  & {\bf 14.7 $\pm$ 1.9} \\
				\bottomrule
			\end{tabular}
	\end{center}
\end{wraptable}
We benchmark the aforementioned imitation learning method on the cartpole swing-up task. We replicate the setup from \cite{gal2016improving}, i.e. pole length $0.6\,\text{m}$, cart mass $0.5\,\text{kg}$, pole mass $0.5\,\text{kg}$, time discretization  $0.1\,\text{s}$, and $p(\bm x_0)=\mathcal{N}(0, 0.04 \bm I)$. The expert dataset consists of 100 trajectories with a length of 40 steps. We evaluate the expert policy as the solution to the $\argmin$-problem (Eq. \ref{eq:control_problem}) via \textit{random shooting}  (RS)  \citep{rao2009survey}. We use a horizon of 10 steps and 1000 particles for RS.
Hence, the trajectories in $\mathcal{D}_{exp.}$ are rather noisy, as shown in Fig. \ref{fig:control_exp}. During training  of  $\text{MPC}_\text{IFT}$ we initially use a prediction horizon of 1 and increase it throughout training. 
We compare our proposed method to \textit{behavioral cloning} (BC) \citep{Bain96aframework}, which learns a policy $\pi:\mathbb{R}^{D_x}\rightarrow \mathbb{R}^{D_u}$. Since we do not observe the control, we map the predicted control directly to the next state via the learnable dynamics function and minimize the MSE between future states. The details of the used network architectures are given in Appx. \ref{app:archtitect_mpc}. 
As shown in table \ref{tab:results_control}, our method $\text{MPC}_\text{IFT}$  outperforms behavioral cloning for 
 and comes with improved generalization capabilities.

\section{Related Work}
\label{sec:intro}

\textit{Recurrent backpropagation} (RBP) \citep{rbp, rbp2} is the first training method for a specific type of implicit neural networks, i.e. infinitely deep recurrent neural networks. Recent work on RBP extended this approach to efficient gradient estimation \citep{liao2018reviving} or scaled it to large neural networks \citep{rnneq, deepeq}. Other lines of work focused on specific network architectures \citep{implicit_dl} or $\argmin$-problem structure, e.g. problems of convex \citep{amos_convex, satnet} or quadratic \citep{optnet, NIPS2017_7132} type. \cite{gould2019deep} and \cite{ zhang2020implicitly} considered constrained non-convex implicit layers as a generic building block. They proposed to evaluate the backward evaluation using the IFT. However, their work used in the implicit layers functions with symbolic second order derivatives \citep{gould2019deep} or estimated explicitly all terms \citep{ zhang2020implicitly}.

\section{Scope and Limitations}
\label{sec:conclusion}

In this work we have introduced the new general purpose framework of \textit{Differentiable Implicit Layers}. 
For the first time implicit layers, without any restriction on problem or solution type, have been scaled to heavily parameterized neural networks
with large output dimensionality. We have demonstrated our framework on a wide scope of applications.
However, our framework assumes that the underlying $\argmin$-problem can be solved accurately. If the solution is incorrect, the Bi-Level IFT (Thm. \ref{thm_biift}) does not apply anymore. It remains open up to which error tolerance convergence of an DIL can be guaranteed.
Preliminary tests suggested a generous tolerance, regarding the error of the $\argmin$-solution. 
Furthermore \textit{Conjugate Gradient} methods with flexible preconditioning \citep{cg_pre, BOUWMEESTER2015276} offer a interesting perspective in order to further speed up and improve the backward evaluation of a DIL.
\section{Acknowledgements}
We thank Michael Tiemann and Katharina Ott for helpful discussions.

\bibliography{newbib}
\small {\bibliographystyle{plainnat}}

\newpage
\appendix
\section{Implicit-Function-Theorem}
\label{app:ift}
\begin{thm}\label{thm_ift} \textbf{(IFT.)}
Let $\bm y$ be the solution to  an parametrized $\argmin$-problem (Eq. \ref{eq:param_argmin}).
The gradient with respect to $\bm x$ (exchangeable $\bm \theta$) is obtained as:
\begin{equation*}
    \frac{d \bm  y}{d \bm x} = -
    \left(\frac{\partial^2 f(\bm y;\bm x, \bm \theta)}{\partial \bm y^2} \right)^{-1}
    \left( \frac{\partial^2 f(\bm y;\bm x, \bm \theta)}{\partial \bm x \partial \bm y}  \right) .
\end{equation*}
\end{thm}

\paragraph{Proof.} 
\begin{align*}
    \frac{\partial f(\bm y;\bm x,  \bm \theta)}{\partial \bm y} & = 0 &&\text{Since $f$ is evaluated at a minimum.}\\
    \frac{d}{d \bm x} \left(\frac{\partial f(\bm y;\bm x,  \bm \theta)}{\partial \bm y} \right) &= 0 &&\text{Differentiate both sides.}\\
    \frac{d}{d \bm x} \left(\frac{\partial f(\bm y;\bm x,  \bm \theta)}{\partial \bm y} \right) &= 
    \frac{\partial^2 f(\bm y;\bm x,  \bm \theta)}{\partial \bm x \partial \bm y}  + 
    \frac{\partial^2 f(\bm y;\bm x,  \bm \theta)}{\partial \bm y^2}\frac{d \bm y}{d\bm x} &&\text{By Chain rule.}\\
    0 &= 
    \frac{\partial^2 f(\bm y;\bm x,  \bm \theta)}{\partial \bm x \partial \bm y}  + 
    \frac{\partial^2 f(\bm y;\bm x,  \bm \theta)}{\partial \bm y^2}\frac{d \bm y}{d\bm x} &&\text{Both results combined.}\\
    \frac{d \bm y}{d\bm x} &= 
    -\left( \frac{\partial^2 f(\bm y;\bm x,  \bm \theta)}{\partial \bm y^2} \right)^{-1}\frac{\partial^2 f(\bm y;\bm x,  \bm \theta)}{\partial \bm x \partial \bm y}   
     &&\text{Final result.}\\
\end{align*}

\subsection{Bi-Level IFT}
\label{app:bilevel}
\paragraph{Theorem \ref{thm_biift} \textbf{(Bi-Level IFT.)}} 
Let $\bm y$ be the solution to  an parametrized $\argmin$-problem (Eq. \ref{eq:param_argmin}).
If $\bm y$ is evaluated on a downstream scalar loss function $l(\bm y)$, the gradient with respect to $\bm x$ (exchangeable $\bm \theta$) is obtained exclusively by 
vector-Matrix products as:
\begin{equation*}
    \frac{d \mathcal{L}}{d \bm x}^T = -
    \underbrace{\frac{\partial \mathcal{L}}{\partial \bm y}^T 
    \overbrace{\left(\frac{\partial^2 f}{\partial \bm y^2} \right)^{-1}}^{\bm H^{-1}}}_{\text{vector-inv. Hessian product $\coloneqq     g$}}
    \left( \frac{\partial^2 f}{\partial \bm x \partial \bm y}  \right)+ {\frac{\partial \mathcal{L}}{\partial \bm x}}^T = -
    \underbrace{\bm g^T \left( \frac{\partial^2 f}{\partial \bm x \partial \bm y}  \right)}_{\text{vector-Jacobian product}}+ {\frac{\partial \mathcal{L}}{\partial \bm x}}^T.
\end{equation*}

\paragraph{Proof.}
\begin{align*}
    \frac{d \mathcal{L}(\bm  y, \bm  x, \bm \theta)}{d \bm x}^T &= \frac{\partial \mathcal{L}(\bm  y, \bm  x, \bm \theta)}{\partial \bm y}^T \frac{d \bm  y}{d \bm x}+ {\frac{\partial \mathcal{L}(\bm  y, \bm  x, \bm \theta)}{\partial \bm x}}^T
    &&\text{Total Derivative}\\
    \frac{d \mathcal{L}(\bm  y, \bm  x, \bm \theta)}{d \bm x}^T &= -\frac{\partial \mathcal{L}(\bm  y, \bm  x, \bm \theta)}{\partial \bm y}^T\left( \frac{\partial^2 f(\bm y;\bm x,  \bm \theta)}{\partial \bm y^2} \right)^{-1}\frac{\partial^2 f(\bm y;\bm x,  \bm \theta)}{\partial \bm x \partial \bm y}+ {\frac{\partial \mathcal{L}(\bm  y, \bm  x, \bm \theta)}{\partial \bm x}}^T   &&\text{$\frac{d \bm y}{d\bm x}$ via IFT.}
\end{align*}

\section{Algorithm for Imitation Learning from Observations with Differentiable MPC}
\label{app:diff_mpc}
\begin{algorithm}[h]
    \scriptsize
	\caption{Imitation Learning from Observations with Differentiable MPC}\label{alg:mpc}
	\begin{algorithmic}
		\State \textbf{Input:} $\mathcal{D}_{exp.}$, True MDP $\bm h_{true}(\cdot)$, Learnable Dynamics $\bm h(\cdot; \bm \theta_h)$, 
		Learnable Cost $ c(\cdot; \bm \theta_c)$
        \Function{\textnormal{\textcolor{black}{\texttt{main}}}}{$\bm x$} 
        \While{not converged}
		\State {\texttt{train\_h}}(\,)
		\State {\texttt{train\_c}}(\,)
		\EndWhile
		\EndFunction %
		
        \Function{\textnormal{\textcolor{black}{\texttt{train\_h}}}}{\,} 
		\State $\bm x \sim \mathcal{ D}_{exp.}$ \Comment{Sample Initial State} 
		\State $\bm u \sim \mathcal{ U}$ \Comment{Sample Random Control} 
		\State $\bm x'=h_{true}(\bm x, \bm u)$ \Comment{Query true MDP}
		\State $\hat {\bm x} = \bm h(\bm u,\bm \theta_h)$ \Comment{Imagine next state}
		\State $\bm \theta_h = \bm \theta_h - \texttt{lr} \nabla_{\bm \theta_h} || \bm x' - \hat {\bm x}||  $	
		\Comment{One Gradient Step on RMSE}
		\EndFunction %
		
		\Function{\textnormal{\textcolor{black}{\texttt{train\_c}}}}{\,} 
		\State $\bm x, \bm x' \sim \mathcal{ D}_{exp.}$ \Comment{Sample subsequent States} 
		\State Set score func. $f = \sum_{t=1}^{{H}} c(\bm u_t, \bm h (\bm x_{t-1}, \bm u_{t-1}; \bm \theta_h); \bm \theta_g)$ 
		\Comment{Initial value $\bm h (\bm x_{0}, \bm u_{0}; \bm \theta_h)=\bm x_{obs}$ (Eq. \ref{eq:control_problem})}
		\State $\bm u_{1:H} = \textcolor{blue}{\texttt{Forward}}(\bm x) $ \Comment{Obtain MPC solution with Alg. \ref{alg:main} via \texttt{RandomShooting}.}
		\State $\hat {\bm x} = \bm h (\bm x, \bm u_{1}; \bm \theta_h)$ \Comment{Execute first control and imagine next state.}
		\State $\bm \theta_c = \bm \theta_c - \texttt{lr} \nabla_{\bm \theta_c} || \bm x' - \hat {\bm x}||  $	
		\Comment{One Gradient Step on RMSE for $\bm \theta_c$. Backward of \textcolor{blue}{\texttt{Forward}} with \textcolor{OliveGreen}{\texttt{Backward}} in Alg. \ref{alg:main}. }
		\EndFunction %
\end{algorithmic}
\end{algorithm}

\section{Differentiable MPC for the Mass-Spring-Damper model}
\label{app:lqr}
\paragraph{Background}  We consider a MPC controller with the cost and policy dynamics obtained by solving an unconstrained infinite-horizon Linear Quadratic Regulator (LQR). The LQR optimizes a quadratic cost function and defines linear dynamics:
\begin{equation}
    \argmin_{u_{0:H}} \sum_{t=0}^{H} \bm x_t^TQ\bm x_t + \bm u_t^TR\bm u_t ~~, s.t.  
    ~\bm x_{t+1} = A \bm x_t + B \bm u_t, ~\bm x_0 = \bm x_{obs.}
    \label{eq:control_problem}
\end{equation}
where $A \in  \mathbb{R}^{n\times n}$ is the state transition matrix, $B \in  \mathbb{R}^{n\times m}$ the input matrix, and $Q  \in  \mathbb{R}^{n\times n}$ and $R  \in  \mathbb{R}^{m\times m}$ are a constant state and  weight matrix respectively. The optimal control action that minimizes Eq. 7 is a linear function of the state and a state feedback gain matrix $K \in  \mathbb{R}^{m\times n}$ \citep{recht2019tour}:
\begin{align}
\bm u_t = -K_t \bm x_t \: ,
\end{align}
for $K$ defined as:
\begin{align}
K = (R + B^TSB)^{-1}B^TSA,
\end{align}
where $S \in  \mathbb{R}^{n\times n}$ satisfies the Discrete Algebraic Ricatti Equation (DARE) : 
\begin{equation}
    \begin{aligned}
A^TSA - S - (A^TSB)(R+B^TSB)^{-1}(B^TSA) + Q = 0.
\end{aligned}
\label{dare_equations}
\end{equation}

As the time horizon tends to infinity the value function and the optimal state feedback gains $K$ are time-invariant. Thus for all $t$ the control can be computed as: $\bm u_t = -K \bm x_t$, which can be obtained as a solution to the DARE. 

In order to use the infinite-horizon LQR in differentiation-based learning, we need to be able to differentiate through the DARE solution. Recently it has been shown how this can be done using an analytic derivative \citep{cannon2020infinite}. Alternatively, we suggest that if we treat the DARE as the optimization problem
\begin{equation}
    \begin{aligned}
\argmin_{S} A^TSA - S - (A^TSB)(R+B^TSB)^{-1}(B^TSA) + Q
\end{aligned}
\label{dare_equations}
\end{equation}
we can use the IFT to compute $\frac{\partial S}{\partial A}$, $\frac{\partial S}{\partial B}$, $\frac{\partial S}{\partial Q}$ and $\frac{\partial S}{\partial R}$. 
For solving the DARE we use build in $\texttt{scipy}$ routines.  

\paragraph{Experimental Setup} The setup is inspired by the imitation learning experiments shown in \cite{cannon2020infinite} and \cite{diffmpc}. The system matrices and initial input are defined as follows:
\[
\mathbf{A} = 
\begin{bmatrix} 
0.00&1.00\\
-\dfrac{k}{m}&-\dfrac{c}{m}
\end{bmatrix},
\mathbf{B} = 
\begin{bmatrix} 0.00 \\ -\dfrac{1}{m} \end{bmatrix},
\mathbf{Q} = 
\begin{bmatrix} 
1.00&0.00\\
0.00&1.00
\end{bmatrix}, 
\mathbf{R} = 2.00, \:
\mathbf{x_0} = 
\begin{bmatrix} 0 \\ 3 \end{bmatrix},
\]
where the state variables ($x_t$) indicate the position and velocity of the given mass $m$. The parameters $k$ and $c$ are a stiffness parameter and a damping coefficient respectively. The values for $m$ and $k$ were fixed to $=1$. The considered $c$ values were $[1, 0.1, -0.6]$. Since the performance was similar for all values, we report results  for $c=1$ only.

The training data is generated by simulating a system for a given $c$ value for the linear system dynamics $\bm x_{t+1}= A \bm x_t + B \bm u_t$. The expert matrix $A$ was used to compute the true control matrix $K$ and the trajectory for $\bm x_t$ was unrolled for a given time horizon. During this process the predicted controls $\bm u_t=-K \bm x_t$ are recorded as the "expert controls" to imitate. The first 50 elements of this trajectory were provided as the training data. At train time a starting point was selected randomly and a prediction 6 steps ahead was made with the current matrix $\hat A$. The learner matrix $\hat A$ was initialized with the correct state transition matrix plus an uniformly distributed random perturbation in the interval $[-0.5, 0.5]$ added to each element. The predicted controls were compared to the experts target controls with the goal to minimize the imitation loss: 
\begin{equation}
  \begin{aligned}
\mathcal{L} = || u_{1:T}(x;A) - u_{1:T}(x;\hat{A}) ||_2^2
\end{aligned} 
\label{imit_loss}
\end{equation}
Note that in contrast to the previous experiment with imitation learning, here the state transitions are not available to the learner.

\begin{figure}[h!]
	\centering
	\begin{subfigure}[t]{0.48\columnwidth}
		\centering\includegraphics[height=0.5\columnwidth]{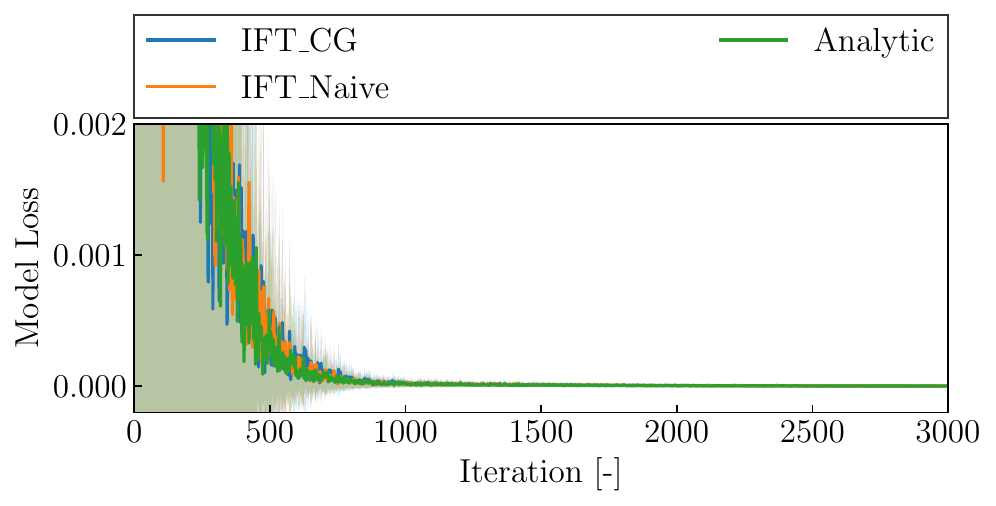}
		\caption{Model Loss.}
		\label{fig:node_times_forward}
	\end{subfigure}%
    \hfill
	\begin{subfigure}[t]{0.48\columnwidth}
	\centering\includegraphics[height=0.5\columnwidth]{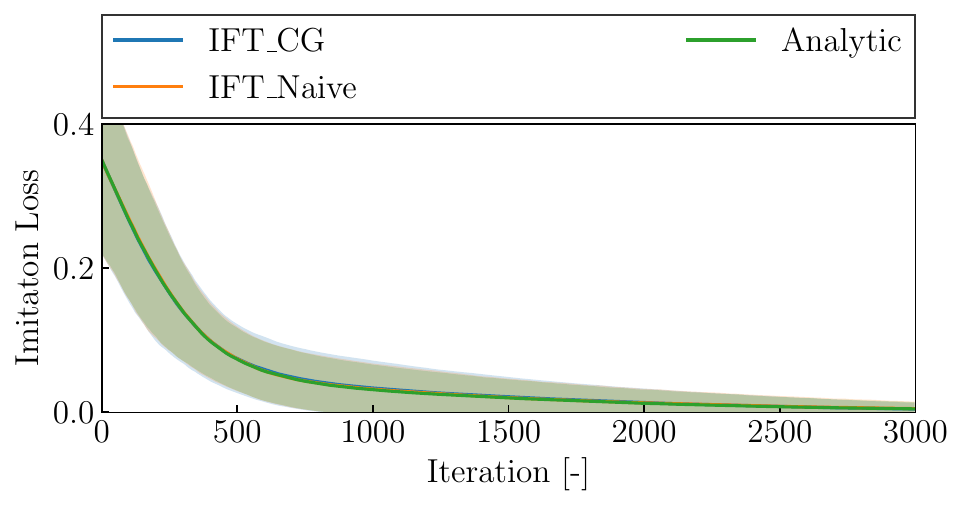}
	\caption{Imitation Loss.}
	\label{fig:node_times_backward}
	\end{subfigure}
	\caption{Average and standard deviation over five different initializations. The imitation loss measures the difference between the expert and learner control values $u$. The model loss is computed as the L2-distance, $||A- \hat A||_2$, between the target expert matrix $A$ and the learner matrix $\hat A$.}
\label{msd_performance_CG_analysis}
\end{figure}

\paragraph{Results} Figure \ref{msd_performance_CG_analysis} shows the imitation and model losses over 3000 optimization iterations. The reported \textit{Analytic} results are obtained by our replication of the analytic gradients, as proposed in \cite{cannon2020infinite}. We can see that for all initializations the imitation loss converges to a low value. Furthermore the declining model loss indicates that the learned dynamics converge to a close approximation of the true dynamics. We can also see that the IFT-CG approach closely follows the performance of the naive implementation, and they show the same learning performance as the analytic gradient. 

\newpage
\section{Architectures}
\label{app:architect}

\subsection{Backward Euler NODE}
\label{app:archtitect_benode}
\paragraph{Van der Pol.}
We use  a single neural network with two hidden layers.
\begin{figure}[H]
\centering
\begin{tikzpicture}[
varnode/.style={rectangle, rounded corners},
layernode/.style={rectangle, draw=blue!60, fill=blue!5, very thick, minimum size=5mm, rounded corners},
]
\node[varnode, node distance=0.5cm]      (input)                              {variable $1\times2$};
\node[layernode, node distance=0.5cm]        (layer1)       [below=of input] {FC-500 + tanh};
\node[layernode, node distance=0.5cm]        (layer2)       [below=of layer1] {FC};
\node[varnode, node distance=0.5cm]        (output)       [below=of layer2] {variable $1\times2$};

\draw[->] (input.south) -- (layer1.north);
\draw[->] (layer1.south) -- (layer2.north);
\draw[->] (layer2.south) -- (output.north);
\end{tikzpicture}
\caption{Neural ODE architecture for VDP.}
\label{node_cmu_architecture}
\end{figure}

\paragraph{Spiral.}
We use  a single neural network with two hidden layers.
\begin{figure}[H]
\centering
\begin{tikzpicture}[
varnode/.style={rectangle, rounded corners},
layernode/.style={rectangle, draw=blue!60, fill=blue!5, very thick, minimum size=5mm, rounded corners},
]
\node[varnode, node distance=0.5cm]      (input)                              {variable $1\times2$};
\node[layernode, node distance=0.5cm]        (layer1)       [below=of input] {FC-50 + tanh};
\node[layernode, node distance=0.5cm]        (layer2)       [below=of layer1] {FC};
\node[varnode, node distance=0.5cm]        (output)       [below=of layer2] {variable $1\times2$};

\draw[->] (input.south) -- (layer1.north);
\draw[->] (layer1.south) -- (layer2.north);
\draw[->] (layer2.south) -- (output.north);
\end{tikzpicture}
\caption{Neural ODE architecture for VDP.}
\label{node_cmu_architecture}
\end{figure}

\newpage
\paragraph{CMU Walking.}
We use a similar architecture as \cite{odevae}.
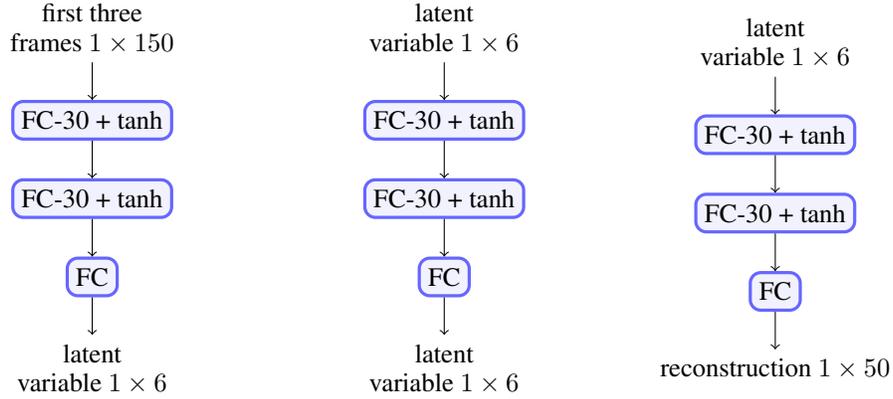
\begin{figure}[H]
\begin{minipage}{.34\textwidth}
\centering
\begin{tikzpicture}[
varnode/.style={rectangle, rounded corners},
layernode/.style={rectangle, draw=blue!60, fill=blue!5, very thick, minimum size=5mm, rounded corners},
]
[align=center,node distance=20cm and 40cm] 
\node[varnode, align=center, ]      (input) {first three \\frames $1\times150$};
\node[layernode, node distance=0.5cm]        (layer1)       [below=of input] {FC-30 + tanh};
\node[layernode, node distance=0.5cm]        (layer2)       [below=of layer1] {FC-30 + tanh};
\node[layernode, node distance=0.5cm]        (layer3)       [below=of layer2] {FC};
\node[varnode, align=center, node distance=0.5cm]        (output)       [below=of layer3] {latent\\variable  $1\times6$};
\draw[->] (input.south) -- (layer1.north);
\draw[->] (layer1.south) -- (layer2.north);
\draw[->] (layer2.south) -- (layer3.north);
\draw[->] (layer3.south) -- (output.north);
\end{tikzpicture}
\end{minipage}%
\begin{minipage}{.33\textwidth}
\centering
\begin{tikzpicture}[
varnode/.style={rectangle, rounded corners},
layernode/.style={rectangle, draw=blue!60, fill=blue!5, very thick, minimum size=5mm, rounded corners},
]
[align=center,node distance=20cm and 40cm] 
\node[varnode, align=center, ]      (input) {latent \\variable $1\times6$};
\node[layernode, node distance=0.5cm]        (layer1)       [below=of input] {FC-30 + tanh};
\node[layernode, node distance=0.5cm]        (layer2)       [below=of layer1] {FC-30 + tanh};
\node[layernode, node distance=0.5cm]        (layer3)       [below=of layer2] {FC};
\node[varnode, align=center, node distance=0.5cm]        (output)       [below=of layer3] {latent\\variable  $1\times6$};
\draw[->] (input.south) -- (layer1.north);
\draw[->] (layer1.south) -- (layer2.north);
\draw[->] (layer2.south) -- (layer3.north);
\draw[->] (layer3.south) -- (output.north);
\end{tikzpicture}
\end{minipage}%
\begin{minipage}{.3\textwidth}
\centering
\begin{tikzpicture}[
varnode/.style={rectangle, rounded corners},
layernode/.style={rectangle, draw=blue!60, fill=blue!5, very thick, minimum size=5mm, rounded corners},
]
[align=center,node distance=20cm and 40cm] 
\node[varnode, align=center, ]      (input) {latent \\variable $1\times6$};
\node[layernode, node distance=0.5cm]        (layer1)       [below=of input] {FC-30 + tanh};
\node[layernode, node distance=0.5cm]        (layer2)       [below=of layer1] {FC-30 + tanh};
\node[layernode, node distance=0.5cm]        (layer3)       [below=of layer2] {FC};
\node[varnode, align=center, node distance=0.5cm]        (output)       [below=of layer3] {reconstruction  $1\times50$};
\draw[->] (input.south) -- (layer1.north);
\draw[->] (layer1.south) -- (layer2.north);
\draw[->] (layer2.south) -- (layer3.north);
\draw[->] (layer3.south) -- (output.north);
\end{tikzpicture}
\end{minipage}
\caption{Encoder-NODE-Decoder neural architectures for CMU.}
\label{node_cmu_architecture}
\end{figure}





\subsection{Differentiable MPC}
\label{app:archtitect_mpc}
Dynamics network architecture is shared across $\text{MPC}_\text{IFT}$ and behavioural cloning. Admissable control
was in the range $[-1, 1]$.
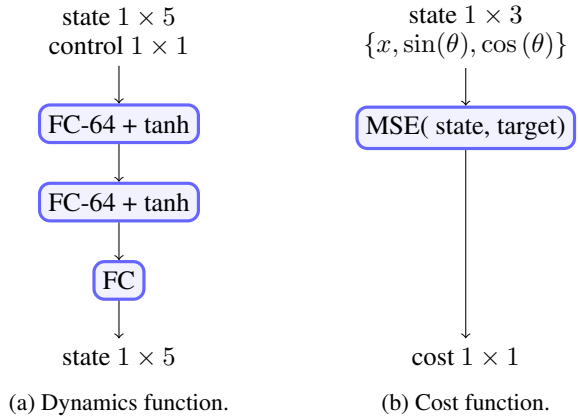
\begin{figure}[h]
\begin{subfigure}[t]{0.33\columnwidth}
\centering
\begin{tikzpicture}[
varnode/.style={rectangle, rounded corners},
layernode/.style={rectangle, draw=blue!60, fill=blue!5, very thick, minimum size=5mm, rounded corners},
]
[align=center,node distance=20cm and 40cm] 
\node[varnode, align=center]      (input) {state $1\times5$\\ control $1\times1$};
\node[layernode, node distance=0.5cm]        (layer1)       [below=of input] {FC-64 + tanh};
\node[layernode, node distance=0.5cm]        (layer2)       [below=of layer1] {FC-64 + tanh};
\node[layernode, node distance=0.5cm]        (layer3)       [below=of layer2] {FC };
\node[varnode, align=center, node distance=0.5cm]        (output)       [below=of layer3] {state $1\times5$};
\draw[->] (input.south) -- (layer1.north);
\draw[->] (layer1.south) -- (layer2.north);
\draw[->] (layer2.south) -- (layer3.north);
\draw[->] (layer3.south) -- (output.north);
\end{tikzpicture}
\caption{Dynamics function.}
\end{subfigure}%
\begin{subfigure}[t]{0.33\columnwidth}
\centering
\begin{tikzpicture}[
varnode/.style={rectangle, rounded corners},
layernode/.style={rectangle, draw=blue!60, fill=blue!5, very thick, minimum size=5mm, rounded corners},
]
[align=center,node distance=20cm and 40cm] 
\node[varnode, align=center]      (input) {state $1\times3$\\ $\{x, \sin(\theta), \cos{(\theta)} \}$};
\node[layernode, node distance=0.5cm]        (layer1)       [below=of input] {MSE( state, target)};
\node[varnode, align=center, node distance=0.5cm]        (output)       [below=of layer3] {cost  $1\times1$};
\draw[->] (input.south) -- (layer1.north);
\draw[->] (layer1.south) -- (output.north);
\end{tikzpicture}
\caption{Cost function.}
\end{subfigure}%

\caption{Architectures used for $\text{MPC}_\text{IFT}$.}
\label{node_cmu_architecture}
\end{figure}

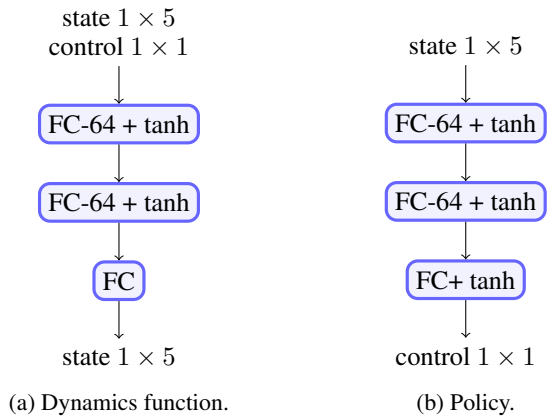
\begin{figure}[h]
\begin{subfigure}[t]{0.33\columnwidth}
\centering
\begin{tikzpicture}[
varnode/.style={rectangle, rounded corners},
layernode/.style={rectangle, draw=blue!60, fill=blue!5, very thick, minimum size=5mm, rounded corners},
]
[align=center,node distance=20cm and 40cm] 
\node[varnode, align=center]      (input) {state $1\times5$\\ control $1\times1$};
\node[layernode, node distance=0.5cm]        (layer1)       [below=of input] {FC-64 + tanh};
\node[layernode, node distance=0.5cm]        (layer2)       [below=of layer1] {FC-64 + tanh};
\node[layernode, node distance=0.5cm]        (layer3)       [below=of layer2] {FC };
\node[varnode, align=center, node distance=0.5cm]        (output)       [below=of layer3] {state $1\times5$};
\draw[->] (input.south) -- (layer1.north);
\draw[->] (layer1.south) -- (layer2.north);
\draw[->] (layer2.south) -- (layer3.north);
\draw[->] (layer3.south) -- (output.north);
\end{tikzpicture}
\caption{Dynamics function.}
\end{subfigure}%
\begin{subfigure}[t]{0.33\columnwidth}
\centering
\begin{tikzpicture}[
varnode/.style={rectangle, rounded corners},
layernode/.style={rectangle, draw=blue!60, fill=blue!5, very thick, minimum size=5mm, rounded corners},
]
[align=center,node distance=20cm and 40cm] 
\node[varnode, align=center]      (input) {state $1\times5$};
\node[layernode, node distance=0.5cm]        (layer1)       [below=of input] {FC-64 + tanh};
\node[layernode, node distance=0.5cm]        (layer2)       [below=of layer1] {FC-64 + tanh};
\node[layernode, node distance=0.5cm]        (layer3)       [below=of layer2] {FC+ tanh};
\node[varnode, align=center, node distance=0.5cm]        (output)       [below=of layer3] {control  $1\times1$};
\draw[->] (input.south) -- (layer1.north);
\draw[->] (layer1.south) -- (layer2.north);
\draw[->] (layer2.south) -- (layer3.north);
\draw[->] (layer3.south) -- (output.north);
\end{tikzpicture}
\caption{Policy.}
\end{subfigure}%
\caption{Architectures used for behavioural cloning.}
\label{node_cmu_architecture}
\end{figure}

\end{document}